\documentclass{article}
\usepackage{ijcai09}

\usepackage{times}

\usepackage{rotating}

\usepackage{latexsym} 

\usepackage{graphicx}

\usepackage{amsmath, amsthm, amssymb}

\usepackage{xspace}

\newcommand{\myOmit}[1]{}

\title{Symmetries of Symmetry Breaking Constraints}

\myOmit{
\author{George Katsirelos \and Toby Walsh}
\institute{NICTA and UNSW, Sydney, Australia\\
\{george.katsirelos,toby.walsh\}@nicta.com.au
}
}

\author{George Katsirelos\\
NICTA, Sydney, Australia\\
george.katsirelos@nicta.com.au
\And
Toby Walsh\\
NICTA and UNSW, Sydney, Australia\\
toby.walsh@nicta.com.au
}

\begin{document}

\maketitle

\begin{abstract}
Symmetry is an important feature of many
constraint programs. We show that \emph{any} symmetry
acting on a set of symmetry breaking constraints 
can be used to break symmetry.
Different symmetries pick out different solutions in each symmetry class. 
We use these observations in two methods
for eliminating symmetry from a problem. 
These methods are designed to have 
many of the advantages of symmetry breaking methods
that post static symmetry breaking constraint
without some of the disadvantages. 
In particular, the two
methods prune the search space using 
fast and efficient propagation of 
posted constraints, whilst
reducing the conflict between symmetry breaking
and branching heuristics. 
Experimental results show that
the two methods perform well on some
standard benchmarks. 
\end{abstract}

\newcommand{\set}{\mathcal}
\newcommand{\myset}[1]{\ensuremath{\mathcal #1}}

\renewcommand{\theenumii}{\alph{enumii}}
\renewcommand{\theenumiii}{\roman{enumiii}}
\newcommand{\figref}[1]{Figure \ref{#1}}
\newcommand{\tref}[1]{Table \ref{#1}}
\newcommand{\myldots}{\ldots}

\newtheorem{mydefinition}{Definition}
\newtheorem{mytheorem}{Observation}
\newtheorem*{myexample}{Running Example}
\newtheorem{mytheorem1}{Theorem}
\newcommand{\myproof}{\noindent {\bf Proof:\ \ }}
\newcommand{\myqed}{\mbox{$\Box$}}
\newcommand{\myend}{\mbox{$\clubsuit$}}

\newcommand{\mymod}{\mbox{\rm mod}}
\newcommand{\mymin}{\mbox{\rm min}}
\newcommand{\mymax}{\mbox{\rm max}}
\newcommand{\range}{\mbox{\sc Range}}
\newcommand{\roots}{\mbox{\sc Roots}}
\newcommand{\myiff}{\mbox{\rm iff}}
\newcommand{\alldifferent}{\mbox{\sc AllDifferent}}
\newcommand{\permutation}{\mbox{\sc Permutation}}
\newcommand{\disjoint}{\mbox{\sc Disjoint}}
\newcommand{\cardpath}{\mbox{\sc CardPath}}
\newcommand{\CARDPATH}{\mbox{\sc CardPath}}
\newcommand{\common}{\mbox{\sc Common}}
\newcommand{\uses}{\mbox{\sc Uses}}
\newcommand{\lex}{\mbox{\sc Lex}}
\newcommand{\usedby}{\mbox{\sc UsedBy}}
\newcommand{\nvalue}{\mbox{\sc NValue}}
\newcommand{\slide}{\mbox{\sc CardPath}}
\newcommand{\sliden}{\mbox{\sc AllPath}}
\newcommand{\SLIDE}{\mbox{\sc CardPath}}
\newcommand{\circularslide}{\mbox{\sc CardPath}_{\rm O}}
\newcommand{\among}{\mbox{\sc Among}}
\newcommand{\mysum}{\mbox{\sc MySum}}
\newcommand{\amongseq}{\mbox{\sc AmongSeq}}
\newcommand{\atmost}{\mbox{\sc AtMost}}
\newcommand{\atleast}{\mbox{\sc AtLeast}}
\newcommand{\element}{\mbox{\sc Element}}
\newcommand{\gcc}{\mbox{\sc Gcc}}
\newcommand{\gsc}{\mbox{\sc Gsc}}
\newcommand{\contiguity}{\mbox{\sc Contiguity}}
\newcommand{\PRECEDENCE}{\mbox{\sc Precedence}}
\newcommand{\assignnvalues}{\mbox{\sc Assign\&NValues}}
\newcommand{\linksettobooleans}{\mbox{\sc LinkSet2Booleans}}
\newcommand{\domain}{\mbox{\sc Domain}}
\newcommand{\symalldiff}{\mbox{\sc SymAllDiff}}
\newcommand{\alldiff}{\mbox{\sc AllDiff}}

\newcommand{\slidingsum}{\mbox{\sc SlidingSum}}
\newcommand{\MaxIndex}{\mbox{\sc MaxIndex}}
\newcommand{\REGULAR}{\mbox{\sc Regular}}
\newcommand{\regular}{\mbox{\sc Regular}}
\newcommand{\precedence}{\mbox{\sc Precedence}}
\newcommand{\STRETCH}{\mbox{\sc Stretch}}
\newcommand{\SLIDEOR}{\mbox{\sc SlideOr}}
\newcommand{\NAE}{\mbox{\sc NotAllEqual}}
\newcommand{\mytheta}{\mbox{$\theta_1$}}
\newcommand{\mysigma}{\mbox{$\sigma_2$}}
\newcommand{\mysigmatwo}{\mbox{$\sigma_1$}}

\newcommand{\todo}[1]{{\tt (... #1 ...)}}

\newcommand{\dpsb}{DPSB}

\section{Introduction}

Symmetry occurs in many problems. 
For instance, certain workers in a staff rostering problem
might have the same skills and availability. If we have a valid schedule, 
we may be able to permute these workers and still have a valid schedule. 
We typically need to factor such symmetry 
out of the search space to be able to find solutions efficiently. 
One popular way to deal with symmetry
is to add constraints which
eliminate symmetric
solutions (see, for instance, 
\cite{puget:Sym,ssat2001,ffhkmpwcp2002,llconstraints06,wecai2006,wcp07,waaai2008}). 
A general method is to
add constraints which limit search to the
lexicographically least solution in each
symmetry class. 
Such symmetry breaking is usually simple to implement 
and is often highly efficient and effective \cite{fhkmwcp2002,fhkmwaij06}.
Even for problems with many symmetries, 
a small number of symmetry breaking 
constraints can often eliminate
much or all of the symmetry. 

One problem with posting symmetry breaking constraints
is that they pick out particular solutions in each
symmetry class, and branching heuristics may conflict
with this choice. In this paper, we consider
two methods for posting symmetry breaking
constraints that tackle this conflict. 
The two methods exploit the observations
that \emph{any} symmetry
acting on a set of symmetry breaking constraints
can be used to break symmetry, and that different
symmetries pick out different solutions. 
The first method 
is {\em model restarts} which
was proposed in \cite{hpsycp08}. 
We periodically restart search with a new model
which contains a different
symmetry of the symmetry breaking constraints. 
The second method posts
a symmetry of the symmetry breaking constraint dynamically during search. 
The symmetry is incrementally chosen to be 
consistent with the branching heuristic.
Our experimental results show that both
methods are effective at reducing the conflict
between branching heuristics and symmetry breaking. 

\section{Background}

A constraint satisfaction problem (CSP) consists of a set of variables,
each with a domain of values, and a set of constraints
specifying allowed combinations of values for subsets of
variables. 
A solution is an assignment 
to the variables satisfying the constraints.
We write $sol(C)$ for the set of all solutions to 
the constraints $C$. 
A common method to 
find a solution of a CSP
is backtracking search. 
Constraint solvers typically prune
the backtracking search space by enforcing a local consistency
property like domain consistency.
A constraint is \emph{domain consistent}
iff for each variable, every value in its domain 
can be extended to an assignment satisfying
the constraint. 
We make a constraint
domain consistent by pruning values for variables which 
cannot be in any satisfying assignment. 
During the search for a solution, a constraint can become
entailed. 
A constraint is \emph{entailed} when
any assignment of values from the respective domains 
satisfies the constraint,
For instance, $X_1<X_n$ is entailed iff
the largest value in the domain of $X_1$ is
smaller than the smallest value in the domain of $X_n$. 
A constraint is 
\emph{dis-entailed} when
its negation is entailed.
For instance, $X<Y$ is dis-entailed if and only if
the smallest value in the domain of $X$ is
larger than or equal to the largest value in the domain of $Y$. 


CSPs can contain symmetry. 
We consider two common types of symmetry
(see \cite{cjjpsconstraints06} for more discussion). 
A \emph{variable symmetry}
is a permutation of the variables
that preserves solutions. 
Formally,
a variable symmetry 
is a bijection $\sigma$ on the
indices of variables such that if $X_1=d_1, \ldots, X_n=d_n$ is a solution
then $X_{\sigma(1)}=d_1, \ldots, X_{\sigma(n)}=d_n$ is also. 
A \emph{value symmetry}, on the other
hand, is a permutation of the values
that preserves solutions. Formally,
a value symmetry 
is a bijection $\theta$ on the
values such that if $X_1=d_1, \ldots, X_n=d_n$ is a solution
then $X_1=\theta(d_1), \ldots, X_n=\theta(d_n)$ is also. 
Symmetries can more generally act 
on both variables and values. 
Our methods also work with such symmetries. 
As the inverse of a symmetry and the identity
mapping are symmetries, the set of symmetries of a problem
forms a group under composition. 
We will use a simple running example
which has a small number of variable
and value symmetries. 
This example demonstrates that we can use
symmetry itself to pick out different solutions
in each symmetry class. 

\begin{myexample}
\renewcommand{\theequation}{\alph{equation}}
The all interval series problem (prob007 in CSPLib.org \cite{csplib})
asks for a permutation 
of 
0 to $n-1$
so that neighbouring differences
form a permutation of 1 to $n-1$. 
We model
this as a CSP 
with
$X_i=j$ iff the $i$th number 
is $j$, and auxiliary variables for
the neighbouring differences. 
One solution for $n=11$ is:
\begin{eqnarray}\label{ais1}
X_1, X_2, \ldots, X_{11} & = & 
3, 7, 4, 6, 5, 0, 10, 1, 9, 2, 8 
\end{eqnarray}
%
This model has a number of different symmetries.
First, there is a variable symmetry $\sigma_{rev}$
that reverses any solution:
\begin{eqnarray} \label{ais2}
X_1, X_2,   \ldots, X_{11} & = & 
 8, 2, 9, 1, 10, 0, 5, 6, 4, 7, 3
\end{eqnarray}
Second, there is a value symmetry $\theta_{inv}$
that inverts values.
If we subtract all values in (\ref{ais1}) from $10$, we
generate a second (but symmetric) solution:
\begin{eqnarray} \label{ais3}
X_1,  X_2,  \ldots, X_{11} & = & 
 7, 3, 6, 4, 5, 10, 0, 9, 1, 8, 2
\end{eqnarray}
Third, we can do both. By reversing and inverting
(\ref{ais1}), we generate a fourth (but symmetric)
solution:
\begin{eqnarray} \label{ais4}
X_1,  X_2,  \ldots, X_{11} & = & 
 2, 8, 1, 9, 0, 10, 5, 4, 6, 3, 7
\end{eqnarray}
The model thus has four symmetries in total: $\sigma_{id}$ (the
identity mapping), $\sigma_{rev}$, $\theta_{inv}$, 
and $\theta_{inv} \circ \sigma_{rev}$. 
\myend 
\end{myexample}

\section{Symmetry breaking}

One common way to deal with symmetry 
is to add constraints to eliminate 
symmetric solutions \cite{puget:Sym}. 
Two important properties of 
symmetry breaking
constraints are soundness and completeness.
A set of symmetry breaking constraint 
is sound iff it leaves at least one solution
in each symmetry class, and complete iff 
it leaves exactly one solution. 

\renewcommand{\theequation}{\arabic{equation}}
\setcounter{equation}{0}

\begin{myexample}
Consider again the all interval
series problem. 
To eliminate the reversal symmetry $\sigma_{rev}$,
we can post the constraint:
\begin{eqnarray} \label{symbreak1}
& X_1 < X_{11}&
\end{eqnarray}
This eliminates solution (\ref{ais2}) as it is 
the reversal of (\ref{ais1}). 
To eliminate the value symmetry $\theta_{inv}$, 
we can post:
\begin{eqnarray} \label{symbreak2}
X_1 \leq 5, & \ \ \ & X_1=5 \Rightarrow X_2 < 5
\end{eqnarray}
This eliminates solution
(\ref{ais3}) as it is the inversion
of (\ref{ais1}). 
Finally, to eliminate the third symmetry $\theta_{inv} \circ \sigma_{rev}$
where we both reverse and invert the solution, 
we can post:
\begin{eqnarray} \label{symbreak3}
\langle X_1, \ldots,  X_6 \rangle &  \leq_{lex} & 
\langle 10-X_{11}, \ldots,  10-X_6 \rangle
\end{eqnarray}
This eliminates solution
(\ref{ais1}) as it is the reversal and inversion
of (\ref{ais4}). 
Note that of the four symmetric solutions given
earlier, only (\ref{ais4})
with $X_1=2$ and $X_{11}=7$
satisfies all these symmetry breaking constraints. 
The other three solutions are eliminated. Thus
(\ref{symbreak1}) to (\ref{symbreak3}) are a sound
and complete set of symmetry breaking constraints. 
\myend \end{myexample}

We now show that {\em any symmetry} 
acting on a set of symmetry breaking constraints
itself breaks the symmetry in a problem. 
Different symmetries pick out different
solutions in each symmetry class. 
To prove this, we need to consider
the action of a symmetry on a symmetry 
breaking constraint. 
Symmetry has been defined acting on assignments. 
We lift this definition to constraints. 
The action of a variable symmetry on
a constraint changes the variables on which the constraint
acts. More precisely, a variable symmetry 
$\sigma$
applied to the constraint 
$C(X_j,\ldots,X_k)$
gives $C(X_{\sigma(j)},\ldots,X_{\sigma(k)})$.
The action of a value symmetry is also easy to
compute. A value symmetry 
$\theta$ applied to the constraint 
$C(X_j,\ldots,X_k)$
gives $C(\theta(X_j),\ldots,\theta(X_k))$. 

\begin{myexample}
To illustrate how we can break symmetry
with the symmetry of a set of symmetry breaking
constraints, we consider symmetries of
(\ref{symbreak1}), (\ref{symbreak2}) and (\ref{symbreak3}).

If we apply $\sigma_{rev}$ to (\ref{symbreak1}), 
we get an ordering constraint that
again breaks the reversal symmetry:
\begin{eqnarray*}
& 
X_{\sigma_{rev}(1)} < X_{\sigma_{rev}(11)}
&
\end{eqnarray*}
This simplifies to: 
\begin{eqnarray*}
& X_{11} < X_{1}&
\end{eqnarray*}
If we apply $\sigma_{rev}$ to (\ref{symbreak2}), 
we get constraints that
again breaks the inversion symmetry:
\begin{eqnarray*} 
X_{11} \leq 5, & \ \ \ & X_{11}=5 \Rightarrow X_{10} < 5
\end{eqnarray*}
Finally, 
if we apply $\sigma_{rev}$ to (\ref{symbreak3}), 
we get a constraint that
again breaks the combined reversal and inversion
symmetry:
\begin{eqnarray*} 
\langle X_{11}, \ldots,  X_6 \rangle &  \leq_{lex} & 
\langle 10-X_{1}, \ldots,  10-X_6 \rangle
\end{eqnarray*}
Note that of the four symmetric solutions given
earlier, only (\ref{ais3})
satisfies 
$\sigma_{rev}$ of
(\ref{symbreak1}), (\ref{symbreak2}) and (\ref{symbreak3}).

We can also break symmetry with any other symmetry of 
the symmetry breaking constraints. 
For instance, 
if we apply $\theta_{inv} \circ \sigma_{rev}$ to (\ref{symbreak1}), 
we get a constraint that
again breaks the reversal symmetry:
\begin{eqnarray*}
& 
10-X_{11} < 10-X_{1}
&
\end{eqnarray*}
This simplifies to: 
\begin{eqnarray*}
& X_{1} < X_{11}&
\end{eqnarray*}
If we apply $\theta_{inv} \circ \sigma_{rev}$ to (\ref{symbreak2}), 
we get constraints that
again breaks the inversion symmetry:
\begin{eqnarray*} 
10- X_{11} \leq 5, & \ \ \ & 10 - X_{11}=5 \Rightarrow 10 - X_{10} < 5
\end{eqnarray*}
This simplifies to:
\begin{eqnarray*} 
X_{11} \geq 5, & \ \ \ & X_{11}=5 \Rightarrow X_{10} > 5
\end{eqnarray*}
Finally, 
if we apply $\theta_{inv} \circ \sigma_{rev}$ to (\ref{symbreak3}), 
we get a constraint that
again breaks the combined reversal and inversion
symmetry:
\begin{eqnarray*} 
\langle 10-X_{11}, \ldots,  10-X_6 \rangle &  \leq_{lex} & 
\langle X_{1}, \ldots,  X_6 \rangle
\end{eqnarray*}
Note that of the four symmetric solutions given
earlier, only (\ref{ais1})
satisfies 
$\theta_{inv} \circ \sigma_{rev}$ of
(\ref{symbreak1}), (\ref{symbreak2}) and (\ref{symbreak3}).
\myend \end{myexample}

The running example illustrates 
that we can break symmetry with a symmetry of 
a set of symmetry breaking constraints. We now
prove that this holds in general: 
\begin{quote}
{\em Any symmetry acting on a set of symmetry
breaking constraints itself breaks
symmetry.}
\end{quote}
More precisely, if a set of 
symmetry breaking constraints is sound,
then any symmetry of these constraints is also sound.
Similarly, if a set of 
symmetry breaking constraints is complete,
then any symmetry of these constraints is also complete. 

\begin{mytheorem}
Given a set of symmetries $\Sigma$ of $C$, 
if $S$ is a sound (complete) set of symmetry breaking
constraints for $\Sigma$
then $\sigma(S)$ for any $\sigma \in \Sigma$
is also a sound (complete) set of symmetry breaking constraints
for $\Sigma$. 
\end{mytheorem}
\myproof
(Soundness)
Consider 
$s \in sol(C \cup S)$. 
Then $s \in sol(C)$
and $s \in sol(S)$. 
Hence $\sigma(s) \in sol(C)$. 
Since $s \in sol(S)$, it follows
that $\sigma(s) \in sol(\sigma(S))$.
Thus, $\sigma(s) \in sol(C \cup \sigma(S))$. 
Hence, there is at least one solution, $\sigma(s)$ 
in every symmetry class of $C \cup \sigma(S)$.
That is, $\sigma(S)$ is a sound set of
symmetry breaking constraints for $\Sigma$. 

(Completeness)
Consider 
$s \in sol(C \cup \sigma(S))$. 
By a similar argument to soundness,
$\sigma^{-1}(s) \in sol(C \cup S)$. 
Hence, there is at most one solution
in every symmetry class of $C \cup \sigma(S)$.
That is, $\sigma(S)$ is a complete set of
symmetry breaking constraints for $\Sigma$. 
\myqed

Different symmetries of the symmetry breaking constraints 
pick out different solutions in each symmetry class.
Thus, if the branching heuristic is going towards a particular
solution, there is a symmetry of the symmetry breaking
constraints which does not conflict with this.

\begin{mytheorem}
Given a symmetry group $\Sigma$,
a sound set $S$ of symmetry breaking
constraints for $\Sigma$,
and any complete assignment $A$, 
then there exists a symmetry $\sigma$ in
$\Sigma$ such that $A$ satisfies $\sigma(S)$. 
\end{mytheorem}
\myproof
Since the set of symmetry breaking constraints
$S$ is sound, it leaves at least one
solution (call it $B$) in the same symmetry class as $A$. That is,
$B$ satisfies $S$. Since
$A$ and $B$ are in the same symmetry class, there exists
a symmetry $\sigma$ in $\Sigma$ with $\sigma(A)=B$.
$\Sigma$ forms a group so also contains the inverse symmetry
$\sigma^{-1}$. Since $B$ satisfies $S$ and $\sigma(A)=B$,
it follows that $\sigma(A)$ satisfies
$S$. Hence $\sigma^{-1}(\sigma(A))$ satisfies $\sigma^{-1}(S)$.
That is, $A$ satisfies $\sigma^{-1}(S)$.
\myqed

\section{Model restarts}

We start with a simple application of these 
observations.
To tackle conflict between branching
heuristics and symmetry breaking constraints,
Heller {\it et al.} propose using {\it model restarts}
\cite{hpsycp08}. In this method, backtracking search is
restarted periodically, using a new model 
which contains different symmetry breaking constraints.
By posting different symmetry breaking 
constraints, we hope at some point
for the branching heuristic and symmetry breaking
not to conflict. 
Our observations that any symmetry
acting on a set of symmetry breaking constraints
can be used to break symmetry, and that different
symmetries pick out different solutions, 
provide us with precisely the tools we need
to perform model restarts to any domain (and not just
to interchangeable variables and values as in \cite{hpsycp08}). 
When we restart
search, we simply post a different symmetry of 
the symmetry breaking constraints. 
We experimented with several possibilities.
The simplest is to choose a symmetry
at random from the symmetry group. 
We also tried various heuristics
like using the symmetry most consistent or
most inconsistent with 
previous choices of the branching heuristic.
However, we observed the best performance of
model restarts with a random choice of symmetry so we 
only report results here with such a choice.

\begin{myexample}
Consider again the all interval series 
problem and posting symmetries of the
symmetry breaking constraints
(\ref{symbreak1}), (\ref{symbreak2}) and (\ref{symbreak3}).
The following table gives the amount of
search needed to find an all interval series of size
$n=11$ using a branching heuristic that
branches in order on the variables introduced
to represent neighbouring differences
in the series, trying values in numerical order. 
This would seem to be a good branching heuristic since
it can find a solution without backtracking. 

{\rm
\begin{center}
\begin{tabular}[h]{|c||r|r|} \hline
Symmetry posted 
& Branches & Time to solve/s \\ 
of  (\ref{symbreak1}) to (\ref{symbreak3})  & & \\ \hline
$\sigma_{id}$ & 1 & 0.00 \\
$\sigma_{rev}$ & 222,758 & 13.74 \\
$\theta_{inv}$ & 425,765 & 24.99 \\
$\theta_{inv} \circ \sigma_{rev}$ & 170,425 & 10.23 \\ \hline
\end{tabular}
\end{center}
}

It is clear from this table that
the different symmetries of the symmetry breaking
constraints interact differently with the branching
heuristic. In particular, the identity symmetry does not
conflict in any way as the branching 
heuristic goes directly to the following solution
at the end of the first branch:
\begin{eqnarray*} 
X_1,  X_2,  \ldots, X_{11} & = & 
0, 10, 1, 9, 2, 8, 3, 7, 4, 6, 5
\end{eqnarray*}
This solution is consistent with 
$\sigma_{id}$ of (\ref{symbreak1}), (\ref{symbreak2})
and (\ref{symbreak3}). 

The other symmetries of the
symmetry breaking constraint
conflict with the branching heuristic. 
In particular, the following symmetry breaking
constraints conflict with this solution:
$\sigma_{rev}$ of (\ref{symbreak1}) as 
$X_{11}=5 > X_1=0$,
$\theta_{inv}$ of (\ref{symbreak1}) 
as 
$10-X_1=10 > 10-X_{11}=5$
and
$\theta_{inv} \circ \sigma_{rev}$ of (\ref{symbreak3})
as $10-X_{11}=5 > X_1=0$. 
As a result, posting these symmetries of the symmetry
breaking constraints increases the search needed
to find a solution. 

Model restarts will help overcome this conflict. 
Suppose we restart search every 100 branches
and choose to post
a random symmetry of (\ref{symbreak1}), (\ref{symbreak2})
and (\ref{symbreak3}).
Let $t$ be the average number of branches
to find a solution. There is $\frac{1}{4}$ chance
that the first restart will post $\sigma_{id}$ of
(\ref{symbreak1}), (\ref{symbreak2})
and (\ref{symbreak3}). In this situation,
we find a solution after 1 branch. 
Otherwise we post
one of the other symmetries of 
(\ref{symbreak1}), (\ref{symbreak2})
and (\ref{symbreak3}). We then
explore 100 branches, reach the cutoff and fail to 
find a solution. As each restart is independent,
we restart and explore on average another $t$ more branches. 
Hence:
\begin{eqnarray*}
t & = & \frac{1}{4} 1 + \frac{3}{4}(100+t)
\end{eqnarray*}
Solving for $t$ gives $t=301$. Thus, using model
restarts with a random symmetry of the symmetry
breaking constraints, we take
just 301 branches on average to find
an all interval series of size $n=11$. 
\myend 
\end{myexample}

Note that posting random symmetries of the symmetry
breaking constraints is not 
equivalent to fixing the symmetry breaking and
randomly branching. As we saw in the example,
different symmetries of the 
symmetry breaking constraints interact in different ways 
with the problem constraints. Although the problem
constraints are themselves initially symmetrical,
branching decisions quickly break the symmetries.

\section{Posting constraints dynamically}

We now consider a more sophisticated
use of the observations that
{any symmetry} 
acting on a set of symmetry breaking constraints
itself breaks symmetry, and
that different symmetries pick out different
solutions in each symmetry class. 
We will incrementally and dynamically post a symmetry of 
the symmetry breaking constraints
which is consistent
with the branching decisions made so far. 
Thus, if the branching heuristic is smart or lucky enough
to branch immediately to a solution,
symmetry breaking will not interfere with this. 

\begin{myexample}
Consider again the all interval
series problem. Suppose we begin by trying
$X_1=10$. Since the $X_i$ are 
all different, $X_{11} \in [0,9]$. Hence, 
the symmetry breaking constraint
$X_{11} < X_1$ is entailed. This is $\sigma_{rev}$ of
(\ref{symbreak1}). It is also $\theta_{inv}$ of
(\ref{symbreak1}). We do not yet need to commit to
which of these two symmetries of the symmetry breaking
constraints we will post. We are sure, however, that
we are not posting $\sigma_{id}$ or
$\theta_{inv} \circ \sigma_{rev}$ of the constraints
(\ref{symbreak1}) to (\ref{symbreak3}).
These two symmetries would require $X_1 > X_{11}$,
and this is dis-entailed. 
We therefore post
$X_{11} < X_1$ and continue search. 
\myend
\end{myexample}

In the example, we posted
symmetry
breaking constraint once they are entailed.
When there are only a few symmetries, we can 
easily implement this
with non-backtrackable variables
and reification. 
Suppose we reify the two ordering constraints:
\begin{eqnarray*}
B_1 \Leftrightarrow (X_1 < X_{11}), & \ \ \ \ \ &
B_2 \Leftrightarrow (X_{11} < X_{1})
\end{eqnarray*}
We then make the Boolean variables, $B_1$ and $B_2$ 
non-backtrackable so that, 
once they are instantiated, their value remains on backtracking. 
We assume that our solver posts the conclusion of
an implication when its hypothesis 
is entailed. Suppose $X_1 < X_{11}$ is
entailed. Then $B_1$ will be set $true$. As $B_1$ is 
non-backtrackable, $X_1 < X_{11}$ will be posted.
Unfortunately, posting symmetry breaking 
constraints like this as soon as they are entailed
may be a little eager. 

\begin{myexample}
Consider again the all interval
series problem. As before, suppose the branching
heuristic has 
set $X_1=10$, and we have posted 
the entailed symmetry breaking constraint
$X_{11} < X_1$. Now $X_1 \geq 5$ 
is also entailed. This is $\theta_{inv}$ of
the first inequality in (\ref{symbreak2}).
If we post this, 
we commit to breaking
symmetry with $\theta_{inv}$ of 
(\ref{symbreak1}) to (\ref{symbreak3}). However, 
this would rule out breaking symmetry with
$\sigma_{rev}$ of (\ref{symbreak1}) to (\ref{symbreak3})
which are also still consistent with the branching
decisions so far. 

Suppose we next branch on $X_{11}=5$. 
The assignments to $X_1$ and $X_{11}$ are only consistent with 
$\theta_{inv}$ of
(\ref{symbreak2}) and of (\ref{symbreak3}).
In fact, both of these 
constraints are now entailed. 
However, $X_1=10$ and $X_{11}=5$ are not consistent
with posting $\sigma_{rev}$ of
(\ref{symbreak3}). This would require that:
\begin{eqnarray*} 
\langle X_{11}, \ldots,  X_6 \rangle &  \leq_{lex} & 
\langle 10-X_{1}, \ldots,  10-X_6 \rangle
\end{eqnarray*}
This is dis-entailed. Hence, our branching
decisions have committed us to break
symmetry with $\theta_{inv}$ of 
(\ref{symbreak1}) to (\ref{symbreak3}).
We therefore post these constraints. 
If search continues, 
we will discover the unique solution
consistent with symmetry breaking and the
initial branching decisions:
\begin{eqnarray*} 
X_1,  X_2,  \ldots, X_{11} & = & 
 10, 0, 9, 1, 8, 2, 7, 3, 6, 4, 5
\end{eqnarray*} \myend
\end{myexample}


There is a tradeoff between posting
symmetry breaking constraints early (so
propagation prunes the search space) and
late (so we do not conflict with future
branching decisions). We propose
the following rule for when to post
symmetry breaking constraints. The 
rule only posts symmetry
breaking constraints once the branching heuristic
has forced their choice. It would, however,
be interesting to explore other more eager or
lazy rules. Suppose $S$ is a set of symmetry
breaking constraints for $\Sigma$, and we have 
posted $T$, a symmetry of a subset of $S$. 
A symmetry $\sigma \in \Sigma$ is \emph{consistent} with $T$ iff
$T$ is entailed by $\sigma(S)$ and \emph{inconsistent} otherwise. 
A symmetry $\sigma \in \Sigma$ is \emph{eliminated} by
posting some symmetry breaking constraint $c$
iff $\sigma$ is consistent with $T$
but inconsistent with $T \cup \{ c \}$. 
The \emph{forced symmetry}
rule is defined as follows:
\begin{quote}
{\em
Given a set of symmetry breaking 
constraints, if during backtracking search a symmetry of one of these 
constraints is entailed, this symmetry is consistent
with previously posted symmetry breaking constraints,
and all symmetries eliminated by this 
entailed constraint are inconsistent
with the current state then 
we post the entailed
constraint. 
}
\end{quote}
We first show that this rule is sound. 

\begin{mytheorem}
Given a set of symmetries $\Sigma$ of $C$, 
if $S$ is a sound set of symmetry breaking
constraints for $\Sigma$
then the forced symmetry rule using $S$ is
a sound symmetry breaking method. 
\end{mytheorem}
\myproof
The rule only permits constraints
of a particular symmetry to be posted. 
By Observation 1, this is sound. 
\myqed

In general, this rule may not be
complete even when given a complete set of 
symmetry breaking constraints.
However, it is easy to modify the rule so that it is 
complete. Whenever we reach a solution, 
we simply pick a consistent
symmetry and post all the symmetry breaking constraints
associated with this symmetry.
We can also define a common property of many 
symmetry breaking constraints
for which the unmodified 
rule is 
complete. 
A set of symmetry breaking constraints
$S$ for the symmetries $\Sigma$ of $C$ is
\emph{proper} iff $S$ is sound and complete for $\Sigma$
and every non-identity symmetry in $\Sigma$ maps 
any solution of $S \cup C$
onto a different solution.
With a proper set of symmetry breaking
constraints, each solution within
a symmetry class is associated with a 
different symmetry.  
For instance, 
constraints (\ref{symbreak1}) to (\ref{symbreak3})
form a proper set of symmetry

\myOmit{
\begin{myexample}
(\ref{symbreak1}), (\ref{symbreak2}) and (\ref{symbreak3})
form a proper set of symmetry
breaking constraints. 
First, they are a sound and complete set. 
Second, consider any of the three non-identity symmetries. 
$\sigma_{rev}$ 
swaps $X_1$ and $X_{11}$. As $X_1 < X_{11}$, $\sigma_{rev}$ maps 
any solution of $S \cup C$ onto a different solution. 
$\theta_{inv}$
maps $X_i$ onto $10-X_i$. There are two cases to consider.
If $X_1 < 5$, then $X_1 \neq \theta_{inv}(X_1)$.
If $X_1=5$, then $X_2 < 5$ and hence $X_2 \neq \theta_{inv}(X_2)$.
In both cases, $\theta_{inv}$ maps any solution 
of $S \cup C$ onto 
a different solution. 
Finally, 
$\theta_{inv} \circ \sigma_{rev}$ 
maps $X_i$ onto $10-X_{12-i}$. Suppose this maps a solution 
of $S \cup C$ onto itself. Now the difference
between $X_{11}$ and $X_{10}$ is $|(10-X_1)-(10-X_2)|$.
This simplifies to $|X_1-X_2|$, the difference
between $X_1$ and $X_2$. But no two differences
are identical. This is a contradiction. 
Hence, the 
action of $\theta_{inv} \circ \sigma_{rev}$ 
must give a different solution.
Thus every non-identity symmetry 
maps a solution of $S \cup C$ onto a different solution. 
Hence this set is proper.
\end{myexample}
}

We new prove that with a proper set of symmetry
breaking constraints, the forced symmetry rule
is a sound \emph{and} complete symmetry breaking method.
That is, it will find exactly one solution in each
symmetry class. 

\begin{mytheorem}
Given a set of symmetries $\Sigma$ of $C$, 
if $S$ is a proper set of symmetry breaking
constraints for $\Sigma$
then the forced symmetry rule is
both sound and complete. 
\end{mytheorem}
\myproof
(Soundness) 
Immediate as a proper set 
is sound.

(Completeness)
Consider the first solution visited.
As the set of symmetry
breaking constraints is proper, 
only one symmetry of these
constraints will be entailed.  All other symmetries
are inconsistent with the current state
and are eliminated. The forced symmetry rule
therefore post this symmetry of the
symmetry breaking constraints. By Observation 1, as the 
symmetry breaking constraints are complete,
this eliminates all other solutions in the same symmetry
class.
\myqed

Finally, we observe that with 
certain symmetry breaking constraints, the forced symmetry
rule is equivalent to 
posting symmetry breaking constraints
as soon as they are entailed. 
For symmetry breaking constraints
like $X_1 < X_{11}$, as soon as
the constraint or its negation is entailed,
all variable symmetries are either consistent or 
they are eliminated.

\section{Interchangeable variables and values}

To test these two symmetry breaking methods,
we consider a common 
type of symmetry where variables and values partition
into interchangeable sets \cite{sellmann2,fpsvcp06}. 
This is sometimes called piecewise variable and value
symmetry. We chose this class of symmetry other 
the many other types of symmetry studied in the past
as it was used in the previous experimental study of model
restarts \cite{hpsycp08}.
Suppose that the $n$ variables partition into $a$ disjoint
sets and variables within each set are interchangeable. 
Similarly, suppose that the $m$ values partition into $b$ disjoint
sets and values within each set are interchangeable. 
We will order variable indices so that $X_{p(i)}$ to $X_{p(i+1)-1}$ is
the $i$th partition of variables for $1 \leq i \leq a$, 
and value indices so that $d_{q(j)}$ to $d_{q(j+1)-1}$ is
the $j$th partition of values for $1 \leq j \leq b$. 

Flener {\it et al.} \cite{fpsvcp06} proved that 
we can eliminate the exponential number of 
symmetries due to such interchangeability
with a polynomial number of symmetry breaking constraints:
\begin{eqnarray*}
& X_{p(i)} \leq \myldots \leq X_{p(i+1)-1} &  \\
& \gcc([X_{p(i)}, \myldots, X_{p(i+1)-1}],[d_1,\myldots,d_m],[O_{1}^i,\myldots,O
_{m}^i]) & \\
 & \hspace{-1em} (O_{q(j)}^1,\myldots,O_{q(j)}^a) \geq_{\rm lex} \myldots \geq_{\rm lex} (O_{q(j+1)-1}^1,\myldots,O_{q(j+1)-1}^a)  & 
\end{eqnarray*}
Where $i \in [1,a]$ and $j \in [1,b]$, and 
$\gcc$ counts
the number of occurrences of the values in each equivalence
class of variables. 
That is, $O_j^i = |\{ k | X_k=d_j, p(i)\leq k < p(i+1)\}|$. 
The {\em signature} of $d_k$
is $(O_{k}^1,\myldots,O_{k}^a)$, the number of 
occurrences of $d_k$ in each variable partition. 
The signature is invariant to the permutation
of variables within each equivalence class. 
By ordering variables within each equivalence class, 
we prevent permutation of interchangeable variables. 
Similarly, by ordering the signatures, we prevent 
permutation of interchangeable values. 

We will post {symmetries} of these
symmetry breaking constraints. 
We consider symmetries that act along two degrees of freedom: 
the order of interchangeable variables within a variable
partition, 
and the order of the signatures of interchangeable values
within a value partition.
Let $\sigma$ be some permutation of the indices of
interchangeable variables.
Then we can break the symmetry of 
variable interchangeability with the following symmetry of
the variable ordering constraints:
\begin{eqnarray*}
& X_{\sigma(p(i))} \leq \myldots \leq X_{\sigma(p(i+1)-1)} & 
\end{eqnarray*}
Similarly let $\theta$ be some permutation of the indices of 
interchangeable values. Then we can break the
symmetry of value interchangeability with this symmetry of
the signature ordering constraints:
\begin{eqnarray*}
& (O_{\theta(q(j))}^1,\myldots,O_{\theta(q(j))}^a) \geq_{\rm lex} \myldots \geq_{\rm lex} (O_{\theta(q(j+1)-1)}^1,\myldots)  & 
\end{eqnarray*}


\myOmit{
\subsubsection{Special case: only interchangeable variables.}

Suppose variables partition into interchangeable
sets but the values do not. Then this method
simplifies to the symmetry breaking rule:
\begin{quote}
{\em If $X_i < X_j$ is entailed, and $X_i$ and $X_j$ are interchangeable then
post 
$X_i \leq X_j$. 
}
\end{quote}

That is, we break symmetry by ordering interchangeable variables
whenever  branching strictly orders them. 

\subsubsection{Special case: only interchangeable values.}

Suppose the values partition into interchangeable sets but
the variables do not. Then 
this method simplifies to a dynamic form
of value precedence \cite{llcp2004} 
in which we order the first occurrence of interchangeable 
values. 
It corresponds to the symmetry breaking rule:
\begin{quote}
{\em If $d_j$ first occurs before $d_k$, and the two
values are interchangeable
then post constraints to ensure
that $d_j$ always first occurs before $d_k$. 
}
\end{quote}
}

\section{Experiments}

\begin{sidewaystable*}[htbp]
  \scriptsize
  \centering
  \begin{tabular}[ht]{|l|r|r|r|r|r|r|r|r|r|r|r|r|r|r|r|r|r|r|}
    \hline
    \# & 
    \multicolumn{6}{|c|}{Static posting} &
    \multicolumn{4}{|c|}{Dynamic posting} &
    \multicolumn{4}{|c|}{SBDS-pair} &
    \multicolumn{4}{|c|}{Model Restarts} \\
    &
    \multicolumn{2}{|c|}{Lex} &
    \multicolumn{2}{|c|}{Antilex} &
    \multicolumn{2}{|c|}{Random} &
    \multicolumn{2}{|c|}{Lex/Antilex} &
    \multicolumn{2}{|c|}{Random} &
    \multicolumn{2}{|c|}{Lex/Antilex} &
    \multicolumn{2}{|c|}{Random} &
    \multicolumn{2}{|c|}{Lex} &
    \multicolumn{2}{|c|}{Random}
    \\
    & 
    opt &
    t /
    b &
    opt &
    t /
    b &
    opt &
    t /
    b &
    opt &
    t /
    b 
    &
    opt &
    t /
    b &
    opt &
    t /
    b 
    & 
    opt &
    t /
    b &
    opt &
    t /
    b &
    opt &
    t /
    b 
    \\
    \hline \hline
    \multicolumn{19}{|c|}{Graph Coloring} \\
    \hline \hline
1& \textbf{13} & \textbf{0.11} &- &- & \textbf{13} &119.64 & \textbf{13} &0.18 & \textbf{13} &1.99 &13 * &0&15 * &0.24 & \textbf{13} &1.14 & \textbf{13} &5.13 \\
&  &387&  &- &  &424 K &  & \textbf{315} &  &4746&  &13253 K &  &14606 K &  &2087&  &13 K \\
2& \textbf{14} &6.29 & \textbf{14} &24.4 & \textbf{14} &16.1 & \textbf{14} &12.36 & \textbf{14} &20.7 &14 * &0.01 &20 * &0.01 & \textbf{14} & \textbf{5.22} & \textbf{14} &23.22 \\
&  &25 K &  &138 K &  &76 K &  &25 K &  &46 K &  &7918 K &  &13719 K &  & \textbf{13 K} &  &54 K \\
3& \textbf{16} & \textbf{0.32} &40 * &1.31 &- &- & \textbf{16} &0.52 &22 * &7.03 &16 * &0.01 &16 * &0.01 & \textbf{16} &27.01 & \textbf{16} &6.18 \\
&  & \textbf{730} &  &2514 K &  &- &  &801&  &1516 K &  &6746 K &  &6492 K &  &38 K &  &12 K \\
5& \textbf{13} &254.93 &- &- &- &- &13 * &119.44 &27 * &0.08 &14 * &0.01 &16 * &0.01 & \textbf{13} & \textbf{117.9} & \textbf{13} &208.4 \\
&  &1001 K &  &- &  &- &  &1368 K &  &1150 K &  &15710 K &  &12505 K &  & \textbf{410 K} &  &693 K \\
6& \textbf{8} & \textbf{0.03} &- &- &- &- & \textbf{8} &0.07 &27 * &131.05 & \textbf{8} &4.13 & \textbf{8} &4.25 & \textbf{8} &0.74 & \textbf{8} &0.77 \\
&  &60&  &- &  &- &  & \textbf{53} &  &1535 K &  &40 K &  &40 K &  &1980&  &1794\\
7& \textbf{17} & \textbf{0.11} &- &- &- &- & \textbf{17} &0.18 &-&-&17 * &0.01 &17 * &0.01 & \textbf{17} &20.67 & \textbf{17} &100.77 \\
&  & \textbf{170} &  &- &  &- &  &185&-&-&  &4697 K &  &4522 K &  &59 K &  &284 K \\
9& \textbf{8 *} & \textbf{0.01} &16 * &445.43 & \textbf{8 *} &25.88 & \textbf{8 *} &0.05 &21 * &170.18 & \textbf{8 *} & \textbf{0.01} & \textbf{8 *} &0.02 & \textbf{8 *} &1.83 & \textbf{8 *} &5.61 \\
&  &3850 K &  &2220 K &  &4550 K &  &1548 K &  &1394 K &  &13795 K &  &13710 K &  &3361 K &  &3463 K \\
10& \textbf{10} & \textbf{0.03} & \textbf{10} &2.17 & \textbf{10} &368.25 & \textbf{10} &0.08 & \textbf{10} &1.77 & \textbf{10} &379.12 & \textbf{10} &385.92 & \textbf{10} &2.75 &- &- \\
&  & \textbf{31} &  &5527&  &1364 K &  &56&  &4219&  &3629 K &  &3629 K &  &6503&  &- \\
11& \textbf{18} & \textbf{166.89} &- &- &- &- & \textbf{18} &353.8 &-&-&18 * &0&18 * &0.01 & \textbf{18} &296.81 &- &- \\
&  & \textbf{511 K} &  &- &  &- &  &772 K &-&-&  &8998 K &  &8691 K &  &619 K &  &- \\
12& \textbf{15} &34.7 &- &- &- &- & \textbf{15} & \textbf{23.91} & \textbf{15} &44.51 &15 * &0.01 &15 * &0.21 &15 * &574.25 & \textbf{15} &436.82 \\
&  &115 K &  &- &  &- &  & \textbf{50 K} &  &91 K &  &5953 K &  &5874 K &  &847 K &  &654 K \\
13&14 * &0.01 &- &- &- &- &14 * &0.04 &27 * &0.04 &14 * &0&14 * &0.02 & \textbf{14} & \textbf{211.28} &- &- \\
&  &1652 K &  &- &  &- &  &1231 K &  &1083 K &  &6680 K &  &6283 K &  & \textbf{528 K} &  &- \\
14&12 * &0.02 &- &- &- &- &12 * &0.04 &25 * &2.58 &12 * &0&12 * &0.02 & \textbf{12} & \textbf{3.56} & \textbf{12} &140.81 \\
&  &2003 K &  &- &  &- &  &1236 K &  &1176 K &  &7754 K &  &6953 K &  & \textbf{7958} &  &386 K \\
15& \textbf{11} & \textbf{0.04} &- &- &- &- & \textbf{11} &0.06 &26 * &396.35 &11 * &0&11 * &0.02 &- &- & \textbf{11} &27.97 \\
&  & \textbf{33} &  &- &  &- &  & \textbf{33} &  &1501 K &  &5483 K &  &5220 K &  &- &  &41 K \\
    \hline 

    \hline \hline
    \multicolumn{19}{|c|}{Concert Hall Scheduling} \\
    \hline \hline
1& \textbf{2894} & \textbf{2.2} & \textbf{2894} &7.56 & \textbf{2894} &2.76 & \textbf{2894} &2.66 & \textbf{2894} &3.88 &1765 * &542.37 &1804 * &568.56 & \textbf{2894} &128.28 & \textbf{2894} &134.08 \\
&  & \textbf{2128} &  &17 K &  &4186&  &2184&  &3733&  &912 K &  &1075 K &  &149 K &  &169 K \\
2& \textbf{2245} & \textbf{1.99} & \textbf{2245} &12.36 & \textbf{2245} &3.79 & \textbf{2245} &3.6 & \textbf{2245} &11.36 &2194 * &417.05 &2194 * &173.82 & \textbf{2245} &136.87 & \textbf{2245} &89.82 \\
&  & \textbf{1836} &  &32 K &  &5716&  &3585&  &14 K &  &1258 K &  &1493 K &  &165 K &  &127 K \\
3& \textbf{2639} & \textbf{20.93} & \textbf{2639} &71.37 & \textbf{2639} &27.92 & \textbf{2639} &25.53 & \textbf{2639} &46.37 &1685 * &167.32 &1873 * &423.7 &2587 * &15.92 &2639 * &192.82 \\
&  & \textbf{15 K} &  &131 K &  &31 K &  &16 K &  &36 K &  &827 K &  &1259 K &  &544 K &  &595 K \\
5& \textbf{3634} & \textbf{3.86} & \textbf{3634} &18.28 & \textbf{3634} &7.9 & \textbf{3634} &4.99 & \textbf{3634} &12.01 &3286 * &108.68 &3487 * &471.78 & \textbf{3634} &166.03 & \textbf{3634} &128.15 \\
&  & \textbf{3797} &  &44 K &  &13 K &  &3930&  &11 K &  &1201 K &  &1543 K &  &206 K &  &171 K \\
7& \textbf{3262} & \textbf{1.46} & \textbf{3262} &8.88 & \textbf{3262} &2.88 & \textbf{3262} &2.59 & \textbf{3262} &6.61 &3224 * &594.55 &3224 * &232.49 & \textbf{3262} &109.35 & \textbf{3262} &114.58 \\
&  & \textbf{1102} &  &19 K &  &3444&  &2315&  &5941&  &1088 K &  &1359 K &  &115 K &  &139 K \\
8& \textbf{3288} & \textbf{3.61} & \textbf{3288} &17.42 & \textbf{3288} &5.22 & \textbf{3288} &4.55 & \textbf{3288} &13.29 &1658 * &475.48 &1725 * &308.22 & \textbf{3288} &199.97 & \textbf{3288} &156.35 \\
&  & \textbf{2606} &  &36 K &  &5781&  &2808&  &12 K &  &928 K &  &1199 K &  &199 K &  &169 K \\
9& \textbf{3434} & \textbf{16.36} & \textbf{3434} &57.81 & \textbf{3434} &24.36 & \textbf{3434} &20.22 & \textbf{3434} &49.16 &2335 * &105.84 &2335 * &43.51 & \textbf{3434} &486.67 &3434 * &112.34 \\
&  & \textbf{14 K} &  &122 K &  &32 K &  &14 K &  &50 K &  &943 K &  &1362 K &  &441 K &  &628 K \\
10& \textbf{2847} & \textbf{4.69} & \textbf{2847} &19.41 & \textbf{2847} &7.52 & \textbf{2847} &6.07 & \textbf{2847} &13.36 &2649 * &153.95 &2647 * &473.14 & \textbf{2847} &231.1 & \textbf{2847} &241.09 \\
&  & \textbf{4888} &  &50 K &  &13 K &  &5030&  &12 K &  &1177 K &  &1626 K &  &297 K &  &332 K \\
11& \textbf{3295} & \textbf{5.31} & \textbf{3295} &33.07 & \textbf{3295} &10.16 & \textbf{3295} &7.91 & \textbf{3295} &34.56 &3295 * &241.42 &3295 * &161.71 & \textbf{3295} &250.43 & \textbf{3295} &261.13 \\
&  & \textbf{3451} &  &62 K &  &12 K &  &5256&  &36 K &  &672 K &  &793 K &  &223 K &  &273 K \\
12& \textbf{1197} & \textbf{11.1} & \textbf{1197} &38.43 & \textbf{1197} &15.05 & \textbf{1197} &13.45 & \textbf{1197} &27.67 &895 * &112.34 &958 * &419.76 & \textbf{1197} &479.43 & \textbf{1197} &241.41 \\
&  & \textbf{10 K} &  &74 K &  &21 K &  &10 K &  &28 K &  &736 K &  &890 K &  &475 K &  &271 K \\
13& \textbf{2565} & \textbf{2.84} & \textbf{2565} &18.03 & \textbf{2565} &5.25 & \textbf{2565} &4.14 & \textbf{2565} &13.45 &2565 * &101.34 &2565 * &173.93 & \textbf{2565} &156.49 & \textbf{2565} &130.58 \\
&  & \textbf{2411} &  &43 K &  &7266&  &3258&  &14 K &  &993 K &  &1187 K &  &186 K &  &170 K \\
14& \textbf{3235} & \textbf{6.91} & \textbf{3235} &25.84 & \textbf{3235} &9.57 & \textbf{3235} &8.21 & \textbf{3235} &15.69 &2385 * &157.02 &2385 * &46.43 & \textbf{3235} &349.81 & \textbf{3235} &373.23 \\
&  & \textbf{5650} &  &50 K &  &12 K &  &5725&  &13 K &  &819 K &  &1168 K &  &324 K &  &398 K \\
15& \textbf{3234} & \textbf{17.95} & \textbf{3234} &63.28 & \textbf{3234} &24.32 & \textbf{3234} &24.71 & \textbf{3234} &44.64 &2168 * &252.34 &2331 * &54.61 &3214 * &449.52 &3234 * &281.55 \\
&  & \textbf{15 K} &  &138 K &  &32 K &  &19 K &  &41 K &  &1122 K &  &1622 K &  &592 K &  &733 K \\
    \hline
  \end{tabular}
  \caption{Static vs Dynamic posting of symmetry breaking constraints on Graph Coloring and Concert Hall Scheduling. ``opt'' is the quality of the solution
found (* indicates optimality was not proved), ``t'' is the runtime in 
seconds, ``b'' is the number of backtracks. The best method for a problem
instance is in {\bf bold font}.}
  \label{tab:results-chs}
\end{sidewaystable*}

We used model restarts and 
the forced symmetry rule to post symmetries dynamically
of the symmetry breaking 
constraints
of Flener {\it et al.} \cite{fpsvcp06}.
Problems are coded into Gecode 2.2.0.
We evaluated the two methods on 
the same two benchmark domains used
in previous studies of symmetry breaking
for interchangeable variables and values
\cite{llwycp07}. 
Experiments
were run on an 2-way Intel Xeon with 6MB of cache and 4 cores in each
processor running at 2GHz.
All instances were terminated after 10 minutes.
We used smallest domain as a variable ordering
heuristic in each experiment. For value
ordering heuristic, we used lexicographical, anti-lexicographical
and random orderings. 

Our experiments are designed to test two hypotheses. 
The first hypothesis is that these two
methods are less prone to conflict between 
branching heuristics and symmetry breaking. 
The second hypothesis is that these two methods
explore a smaller search tree than dynamic methods
like SBDS. This is due to both the propagation of the posted
symmetry breaking constraints and the need to limit
SBDS to just generators to make it computationally 
tractable. 
We limit our comparison of dynamic methods
to comparison against SBDS. Whilst there is a specialized
dynamic symmetry breaking
method for interchangeable variables and
values, experiments in \cite{hpsycp08}
show that this is several orders of magnitude
slower than static methods. 
In addition, dominance detection methods
like SBDD are shown to be three orders
of magnitude slower than static methods in 
\cite{hpsycp08}. 
Finally, we used SBDS to break
just generators of the symmetry group
as breaking the full symmetry group 
quickly ran out of memory. 
We used SBDS with two different sets
of generators: one set has a generator
that exchanges each
pair of consecutive variables/values
in each partition; the other has a
generator that exchanges
the first two variables/values in each partition
and one that rotate all variables/values.
We got similar results with both and
thus here report only results for
the first set, denoted SBDS-pair in
Table 1.

The first set of experiments uses random graph coloring
problems generated in the same way as the previous 
experimental study in \cite{llwycp07}. 
There is a variable for each vertex
and not-equals constraints between variables corresponding
to connected vertices. 
All values in this
model are interchangeable. 
In addition, we introduce variable symmetry 
by partitioning variables into 
interchangeable sets of size at most 8. We
randomly connect the vertices within each partition with either a
complete graph or an empty graph, and choose each option with equal
probability. Similarly, between any two partitions there is equal
probability that the partitions are completely connected or
independent. Results for graphs with 40
vertices are shown in the top half of Table 1. 

The second set of experiments uses a more structured benchmark which
is again taken from a previous 
experimental study \cite{llwycp07}. 
In the concert hall scheduling
problem, we have $n$ applications to use one of $m$ identical
concert halls. Each application has a start and end time as well as an
offer for the hall. We accept applications so that their intervals do
not overlap and the profit (the sum of the offers of accepted
applications) is maximized. We randomly generate instances so that
applications are split into partitions of size at most 8 and within
each partition all applications have the same start and end time and
offer. Our model 
assigns $X_i=j$ if the $i^{th}$
application is accepted and placed in hall $j$, and $X_i=m+1$ if it is rejected.
Variables corresponding to applications in the same
partition are interchangeable. Values divide into two
partitions: the values $1$ to $m$ are interchangeable, while the value
$m+1$ is in a separate partition. Results for 
instances with 40 applications and 10 
halls are
shown in the bottom half of Table~1. 

The results support both our hypotheses. Both
methods are less prone to conflict between symmetry breaking
and the branching heuristic. 
With both SBDS and our forced symmetry rule for
dynamically posting symmetry breaking constraints,
we obtained the same results with the lexicographical
and the (inverse) anti-lexicographical value ordering
heuristic. 
With model restarts, results 
with the lexicographical and the anti-lexicographical value ordering
heuristic are sufficiently similar that we only report
the former. Our second hypothesis, that the two methods explore a
smaller search tree than SBDS is also confirmed. SBDS was unable to
prove optimality in all but one instance.
In addition, on the harder 
graph coloring instances, both methods tend to outperform the static
method. It is hard, however, to choose between the two 
methods. The model restarts method offers slightly better performance on the
graph coloring instances, whilst our method of dynamically posting
static symmetry breaking constraints offers better
performance on the concert hall scheduling instances. 

\section{Other related work}

Closest in spirit to our forced symmetry rule for
dynamically posting symmetry breaking constraints is SBDS
\cite{backofen:sym,sbds,bwconstraints02}.
SBDS can work
with any type of branching decision but for simplicity
we assume that branching decisions are of the form $Var=val$.
All current implementations of SBDS make this assumption. 
If we have a symmetry $\sigma$, 
the partial assignment $A$ and 
have explored and rejected $Var=val$ then on
backtracking, SBDS posts:
$$ \sigma(A) \rightarrow \sigma(Var \neq val)$$
This ensures that we never explore the symmetric
state to the one that has just been excluded. 
Our forced symmetry rule
also posts static symmetry breaking dynamically
during search. However, 
the two methods differ along three important dimensions.
First, SBDS posts symmetry breaking 
constraints when backtracking and exploring
the second branch of the search tree; the forced symmetry rule, on the
other hand, can post symmetry breaking constraints
down either branch. 
Second, SBDS posts symmetries of the current nogood;
the forced symmetry rule, on the
other hand, can post {\em any} type of symmetry breaking constraint.
Here, for instance, it posts ordering constraints
on the signatures. 
Third, whilst neither method conflicts
with the branching heuristic if the branching
heuristic goes directly to a solution, 
the forced symmetry rule may conflict with 
the branching heuristic later in search. 
Constraint propagation on constraints
posted by the forced symmetry rule can 
prune values that branching might have taken. 

There are a number of other related methods.
Jefferson {\it et al,} have
proposed GAPLex, a hybrid method
that also combines together static and dynamic symmetry
breaking \cite{gaplex}. 
However, 
GAPLex is limited to dynamically posting 
lexicographical ordering constraints, and to 
searching with a fixed variable ordering. As a consequence, 
GAPLex performs poorly when there are large numbers of 
symmetries. In addition, 
GAPLex is unable to profit from effective dynamic variable ordering
heuristics. 
Puget has also proposed ``Dynamic Lex'', a hybrid method that
dynamically posts static symmetry breaking
constraints during search which works with dynamic
variable ordering heuristics \cite{puget2003}.
This method adds lexicographical ordering symmetry
breaking constraints dynamically during search that
are compatible with the current partial assignment. 
In this way, the first solution found during tree search
is not pruned by symmetry breaking. Unfortunately
Dynamic Lex needs to compute the stabilizers of the current partial
assignment. This requires a potentially expensive
graph isomorphism problem to be solved at each node of the search tree. 
Whilst Dynamic Lex works with dynamic variable
ordering heuristics, it assumes that
values are tried in order. 
Finally Dynamic Lex is limited to posting
lexicographical ordering constraints. This
is problematic when there are many symmetries. 
A direct comparison of our methods
with Dynamic Lex would be interesting but
poses some challenges. 
For instance, Heller {\it et al.}  \cite{hpsycp08} do not compare
model restarts with Dynamic Lex, arguing:
\begin{quote}
{\em ``It is not clear how this method [Dynamic Lex] can be generalized, 
though, and for the case
of piecewise variable and value symmetry, no method with similar 
properties is known yet.''}
\end{quote}

\section{Conclusions}

We proved that {any symmetry} 
acting on a set of symmetry breaking constraints
itself breaks symmetry, and
that different symmetries pick out different
solutions in each symmetry class. 
These observations can be used to
reduce the conflict between symmetry breaking 
and branching heuristics.
We have studied two methods
for breaking symmetry that tackle
this conflict. The first method uses {\em model restarts} which
was proposed in \cite{hpsycp08}. 
We periodically restart search with a new model
which contains a random
symmetry of the symmetry breaking constraints. 
The second method posts
a symmetry of the symmetry breaking constraint dynamically during search. 
The symmetry is incrementally chosen to be 
consistent with the branching heuristic.
The two methods benefit from propagation of 
the posted symmetry breaking constraints, whilst
reducing the conflict between symmetry breaking
and branching heuristics. 
Experimental results demonstrated that
the two methods perform well on some
standard benchmarks.

\section*{Acknowledgements}

This research is funded by the Department of Broadband, Communications
and the Digital Economy, and the ARC through Backing Australia’s Ability and
the ICT Center of Excellence program.

\bibliographystyle{named}
\bibliography{/home/tw/biblio/a-z,/home/tw/biblio/a-z2,/home/tw/biblio/pub,/home/tw/biblio/pub2}

\begin{thebibliography}{10}

\bibitem{puget:Sym}
Puget, J.F.:
\newblock On the satisfiability of symmetrical constrained satisfaction
  problems.
\newblock In Komorowski, J., Ras, Z., eds.: Proceedings of ISMIS'93. LNAI 689,
  Springer-Verlag (1993)  350--361

\bibitem{ssat2001}
Shlyakhter, I.:
\newblock Generating effective symmetry-breaking predicates for search
  problems.
\newblock In: Proceedings of LICS workshop on Theory and Applications of
  Satisfiability Testing (SAT 2001). (2001)

\bibitem{ffhkmpwcp2002}
Flener, P., Frisch, A., Hnich, B., Kiziltan, Z., Miguel, I., Pearson, J.,
  Walsh, T.:
\newblock Breaking row and column symmetry in matrix models.
\newblock In: 8th International Conference on Principles and Practices of
  Constraint Programming (CP-2002), Springer (2002)

\bibitem{llconstraints06}
Law, Y., Lee, J.:
\newblock {Symmetry Breaking Constraints for Value Symmetries in Constraint
  Satisfaction}.
\newblock Constraints \textbf{11}(2--3) (2006)  221--267

\bibitem{wecai2006}
Walsh, T.:
\newblock Symmetry breaking using value precedence.
\newblock In: Proceedings of the 17th ECAI, European Conference on Artificial
  Intelligence, IOS Press (2006)

\bibitem{wcp07}
Walsh, T.:
\newblock Breaking value symmetry.
\newblock In: 13th International Conference on Principles and Practices of
  Constraint Programming (CP-2007), Springer-Verlag (2007)

\bibitem{waaai2008}
Walsh, T.:
\newblock Breaking value symmetry.
\newblock In: Proceedings of the 23rd National Conference on AI, Association
  for Advancement of Artificial Intelligence (2008)

\bibitem{fhkmwcp2002}
Frisch, A., Hnich, B., Kiziltan, Z., Miguel, I., Walsh, T.:
\newblock Global constraints for lexicographic orderings.
\newblock In: 8th International Conference on Principles and Practices of
  Constraint Programming (CP-2002), Springer (2002)

\bibitem{fhkmwaij06}
Frisch, A., Hnich, B., Kiziltan, Z., Miguel, I., Walsh, T.:
\newblock Propagation algorithms for lexicographic ordering constraints.
\newblock Artificial Intelligence \textbf{170}(10) (2006)  803--908

\bibitem{hpsycp08}
Heller, D., Panda, A., Sellmann, M., Yip, J.:
\newblock Model restarts for structural symmetry breaking.
\newblock In: 14th International Conference on the Principles and Practice of
  Constraint Programming. (2008)  539--544

\bibitem{cjjpsconstraints06}
Cohen, D., Jeavons, P., Jefferson, C., Petrie, K., Smith, B.:
\newblock Symmetry definitions for constraint satisfaction problems.
\newblock Constraints \textbf{11}(2--3) (2006)  115--137

\bibitem{csplib}
Gent, I., Walsh, T.:
\newblock {CSPLib}: a benchmark library for constraints.
\newblock Technical report, Technical report APES-09-1999 (1999) A shorter
  version appears in the Proceedings of the 5th International Conference on
  Principles and Practices of Constraint Programming (CP-99).

\bibitem{sellmann2}
Sellmann, M., Hentenryck, P.V.:
\newblock Structural symmetry breaking.
\newblock In: Proceedings of 19th IJCAI, International Joint Conference on
  Artificial Intelligence (2005)

\bibitem{fpsvcp06}
Flener, P., Pearson, J., Sellmann, M., Hentenryck, P.V.:
\newblock Static and dynamic structural symmetry breaking.
\newblock In: Proceedings of 12th International Conference on Principles and
  Practice of Constraint Programming (CP2006), Springer (2006)

\bibitem{llwycp07}
Law, Y.C., Lee, J., Walsh, T., Yip, J.:
\newblock Breaking symmetry of interchangeable variables and values.
\newblock In: 13th International Conference on Principles and Practices of
  Constraint Programming (CP-2007), Springer-Verlag (2007)

\bibitem{backofen:sym}
Backofen, R., Will, S.:
\newblock Excluding symmetries in constraint-based search.
\newblock In Jaffar, J., ed.: Proceedings of the 5th International Conference
  on Principles and Practice of Constraint Programming. Number 1713 in Lecture
  Notes in Computer Science, Springer-Verlag (1999)  73--87

\bibitem{sbds}
Gent, I., Smith, B.:
\newblock Symmetry breaking in constraint programming.
\newblock In Horn, W., ed.: Proceedings of ECAI-2000, IOS Press (2000)
  599--603

\bibitem{bwconstraints02}
Backofen, R., Will, S.:
\newblock Excluding symmetries in constraint-based search.
\newblock Constraints \textbf{7}(3-4) (2002)  333--349

\bibitem{gaplex}
Jefferson, C., Kelsey, T., Linton, S., Petrie, K.:
\newblock Gaplex: Generalised static symmetry breaking.
\newblock In: Proceedings of 6th International Workshop on Symmetry in
  Constraint Satisfaction Problems (SymCon-06), held alongside CP-06. (2006)

\bibitem{puget2003}
Puget, J.F.:
\newblock Symmetry breaking using stabilizers.
\newblock In Rossi, F., ed.: Proceedings of 9th International Conference on
  Principles and Practice of Constraint Programming (CP2003), Springer (2003)

\end{thebibliography}

\end{document}